%% file: main_v5.tex
\documentclass[10pt,twocolumn,letterpaper]{article}

\usepackage{cvpr}
\usepackage{times}
\usepackage{epsfig}
\usepackage{graphicx}
\usepackage{amsmath}
\usepackage{amssymb}
\usepackage{verbatim}

\usepackage{multirow}
\usepackage{caption}
\usepackage{subcaption}
\usepackage{algorithm}
\usepackage{algpseudocode}
\usepackage{float}
\usepackage{booktabs}

\usepackage{color}
\usepackage{framed}
\usepackage{tabularx}
\usepackage{threeparttable}

\usepackage[pagebackref=true,breaklinks=true,letterpaper=true,colorlinks=true,bookmarks=false,hyperfootnotes=true,linkcolor=red]{hyperref}

 \def\vs{\emph{vs. }}

\cvprfinalcopy 

\def\cvprPaperID{1176} 

\ifcvprfinal\fi
\begin{document}

\title{\textit{Not All Pixels Are Equal}: Difficulty-Aware Semantic Segmentation \\ via Deep Layer Cascade}

\author{Xiaoxiao Li$^{1}$ \hspace{9pt} Ziwei Liu$^{1}$ \hspace{9pt} Ping Luo$^{2,1}$ \hspace{9pt} Chen Change Loy$^{1,2}$ \hspace{8pt} Xiaoou Tang$^{1,2}$\\
$^1$\small{Department of Information Engineering, The Chinese University of Hong Kong}\\
$^2$\small{Shenzhen Key Lab of Comp. Vis. \& Pat. Rec., Shenzhen Institutes of Advanced Technology, CAS, China}\\
{\tt\small \{lx015,lz013,pluo,ccloy,xtang\}@ie.cuhk.edu.hk}
}

\maketitle

\begin{abstract}
\input{abstract_v4_ccloy_pluo.tex}
\end{abstract}

\section{Introduction}
\label{sec:intro}

\input{intro_v6_ccloy_pluo.tex}
\section{Related Work}
\label{sec:related_work}

\input{relatedwork_v5_ccloy_zwliu.tex}

\section{Deep Layer Cascade (LC)}
\label{sec:methodology}

\input{approach_v9_xxli_zwliu.tex}

\section{Experiments}
\label{sec:experiments}

\input{experiments_v8_cavan_zwliu.tex}

\section{Conclusion}

\input{conclusion_v3_ccloy.tex}

{\small
\bibliographystyle{ieee}
\bibliography{short,egbib}
}

\end{document}

%% file: abstract_v4_ccloy_pluo.tex
We propose a novel deep layer cascade (LC) method to improve the accuracy and speed of semantic segmentation. Unlike the conventional model cascade (MC) that is composed of multiple independent models, LC treats a single deep model as a cascade of several sub-models.
Earlier sub-models are trained to handle easy and confident regions, and they progressively feed-forward harder regions to the next sub-model for processing.
%
Convolutions are only calculated on these regions to reduce computations.
The proposed method possesses several advantages.
First, LC classifies most of the easy regions in the shallow stage and makes deeper stage focuses on a few hard regions. Such an adaptive and `difficulty-aware' learning
improves segmentation performance.
Second, LC accelerates both training and testing of deep network thanks to early decisions in the shallow stage.
Third, in comparison to MC, LC is an end-to-end trainable framework, allowing joint learning of all sub-models.
We evaluate our method on PASCAL VOC and Cityscapes datasets, achieving state-of-the-art performance and fast speed. 

%% file: intro_v6_ccloy_pluo.tex
Semantic image segmentation enjoys wide applications, such as video surveillance \cite{farabet2013learning, TSN2016ECCV} and autonomous driving \cite{geiger2012we, Cordts2016Cityscapes}.
Recent advanced deep architectures, such as the residual network (ResNet) \cite{He2015} and Inception \cite{szegedy2016inception}, significantly improve the accuracy of image segmentation by increasing the depth and number of parameters in deep models.
For example, ResNet-101 is six times deeper than VGG-16 \cite{simonyan2014very} network, with the former outperforms the latter by 4 percent on the challenging PASCAL VOC 2012 image segmentation benchmark \cite{everingham2010pascal}.

\begin{figure}[t]
\begin{center}
\includegraphics[width=0.48\textwidth]{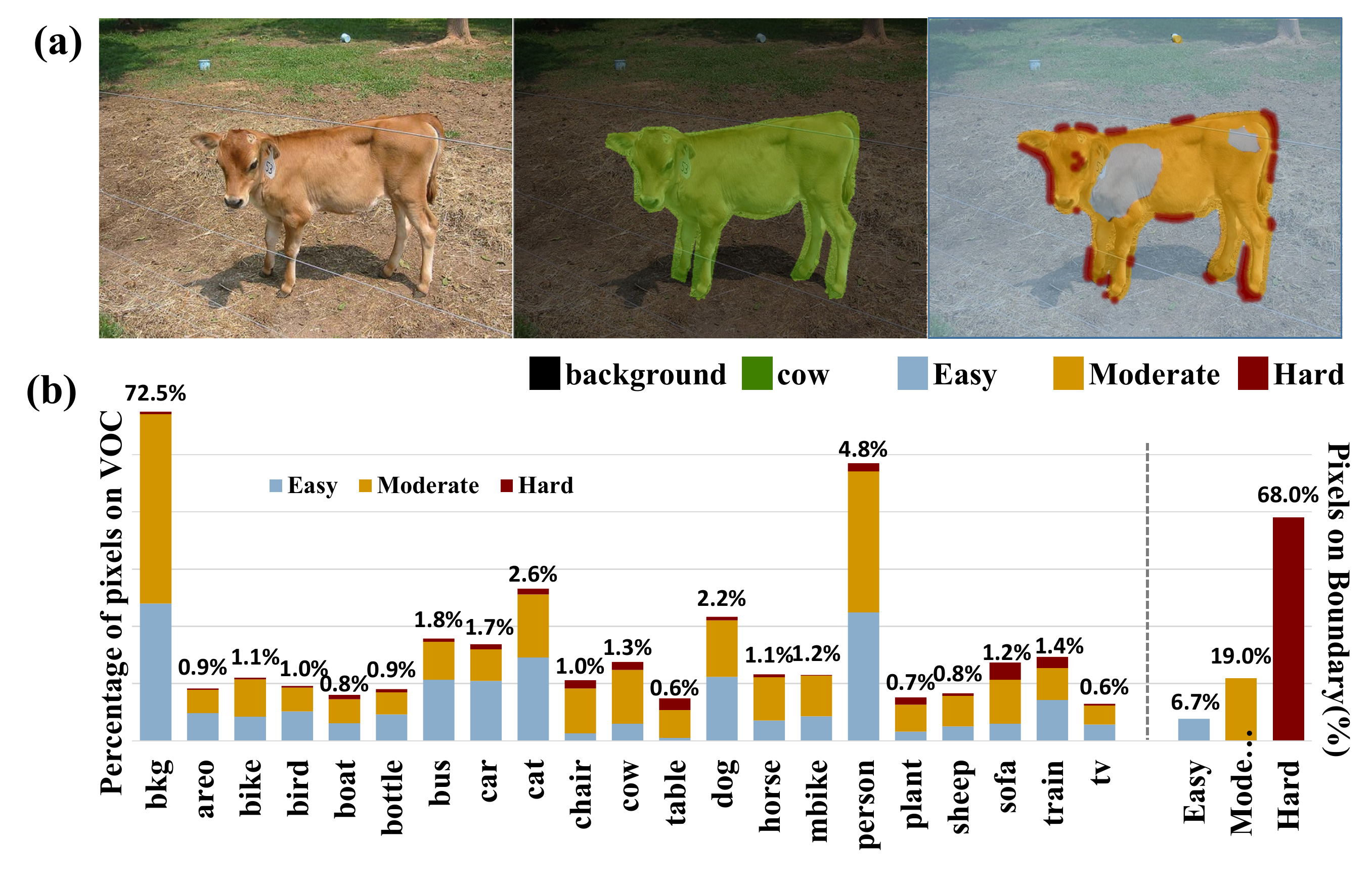}
\vskip -0.2cm
\caption{\small{(a) shows an image of `cow' and `background' (left) and its ground truth label map (middle) from the Pascal VOC 2012 dataset. The difficulty level (\eg recognizability) of pixels are visualized in the right image, where pixels are partitioned into three sets, including `easy' (ES), `moderate' (MS), and `extremely hard' (HS) sets. (b) depicts two histograms. The left one plots the percentages of pixels in VOC validation set with respect to each object category. It can be observed that ES occupies at least 30\% pixels of most objects. The right one reveals that 70\% pixels in HS are located at object boundaries, which have large ambiguity. 
\textbf{Best viewed in color with 300\% zoom}.}}
\label{fig:intro}
\vspace{-20pt}
\end{center}
\end{figure}

Although promising results can be achieved through the increase of model capacity, they come with a price of runtime complexity,
%
%
which impedes the deployments of existing deep models in many applications that demand real-time performance.
For instance, the segmentation speeds of VGG, ResNet-101, and Inception-ResNet on a 300$\times$500 image are 5.7, 7.1 and 9.0 frame per second (FPS), which are far away from real time.
To address this issue, this work presents \emph{Deep Layer Cascade} (LC), which not only substantially reduces the runtime of deep models, but also improves their segmentation accuracies.
%
%
Many deep architectures, including VGG, ResNet, and Inception, can benefit from the above appealing properties by adapting their structures into LC.

Layer Cascade inherits the advantage of the conventional model cascade (MC) \cite{li2015convolutional, viola2001rapid}, which has multiple stages and usually trains one classifier in each stage.
MC is capable of increasing both speed and accuracy for object detection, because
the earlier stages (classifiers) reject most of the easy samples (detection windows) and the later stages can pay attention on a small number of difficult samples, thus reducing false alarms.
Different from MC, LC is carefully devised for deep models in the task of image segmentation.
It considers different layers in a deep network as different stages. In particular, most of the pixels in an image are recognizable by the lower stages and the higher stages, which typically possess far more parameters than the bottom layers, are learned to recognize a small set of challenging pixels.
In this case, the runtime of deep models can be significantly reduced by LC.
Moreover, unlike MC that learns the current stage by keeping all previous stages fixed, LC trains all stages jointly to boost performance.
%

%

Another important difference between LC and MC is the cascade strategy.
In MC, the current stage propagates a sample to the next stage, if its classification score or probability (\ie the response after softmax) is higher than a large threshold, such as $0.95$, indicating that this sample is classified as positive by the current stage with 95\% confidence.
In other words, later stages refine the labels of samples that are considered highly positive in the previous stages, so as to reduce false alarms.

In contrast, LC `rejects' samples with high scores in earlier stages, but those samples with low and moderate confidences are propagated forward.
%
Figure~\ref{fig:intro} takes the segmentation results of LC as an example to illustrate this cascade strategy.
In (a), an image of `cow' and `background' and its ground truth label map from the VOC validation set (VOC val) are shown on the left and middle respectively.
We partition all pixels in the validation set into three different sets, namely ``easy'', ``moderate'', and ``extremely hard'' sets.
The easy set (ES) contains pixels that are correctly classified with larger than 95\% confidence, while the extremely hard set (HS) comprises pixels that are misclassified with larger than 95\% confidence.
The moderate set (MS) covers pixels that have classification scores smaller than 0.95.

In a certain stage of LC, ES and HS are discarded and MS is propagated to the next stage, because of the following two reasons.
First, as shown in the right histogram of Fig.~\ref{fig:intro}(b), we observe that almost 70 percent\footnote{We found that the other 30 percent pixels in HS have wrong annotations. Since our purpose is to improve speed and accuracy of deep models, we do not correct those wrong annotations to enable a fair comparison with previous works.} pixels in HS are located on the boundaries between objects, demonstrating that these pixels are extremely hard to be recognized because of large ambiguity.
An example is given by the right image of Fig.~\ref{fig:intro}(a).
Fitting HS during training may lead to over-fitting in the test stage.
Second, the left histogram of Fig.~\ref{fig:intro}(b) plots the percentages of pixels with respect to each object category in VOC val.
For most of the categories, we found that at least 30 percent pixels belong to ES.
As the background pixels are dominated (72.5\%), rejecting ES and HS reduces more than 40 present pixels in earlier stages and thus significantly reduces computations of deep networks, while improves accuracy, by
enabling deeper layers to focus on foreground objects.

This study makes three main \textbf{contributions}.
(1) This is the first attempt to identify the segmentation difficulty of pixels for deep models.
With this observation, a novel \emph{Deep Layer Cascade} (LC) approach is proposed to significantly reduce computations of deep networks while improving their segmentation accuracies.
%
%
(2) LC's 
properties can be easily applied to many recent advanced network structures.
%
%
After applying LC on Inception-ResNet-v2 (IRNet)~\cite{szegedy2016inception}, its speed and accuracy are improved by 42.8\% and 1.7\%, respectively.
(3) Connections between LC and previous models such as model cascade, deeply supervised network \cite{lee2015deeply}, and dropout \cite{srivastava2014dropout} are clearly presented.
Extensive studies are conducted to demonstrate the superiority of LC.

%% file: relatedwork_v5_ccloy_zwliu.tex
\begin{figure*}[t]
	\centering
	\includegraphics[width=1.0\textwidth]{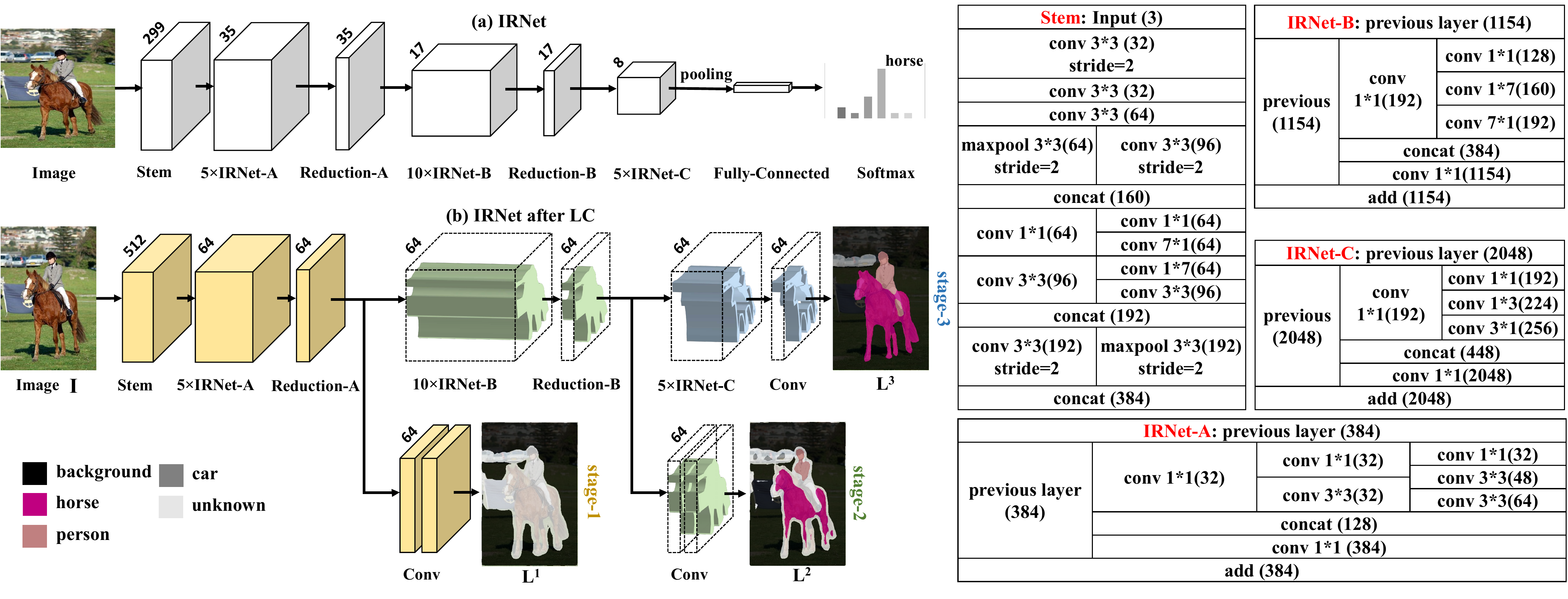}
	\vskip -0.3cm
	\caption{\small{(a) depicts the Inception-ResNet-v2 (IRNet) for classification task. (b) is the architecture of Layer Cascade IRNet (IRNet-LC). The tables at the right show the structure of IRNet.}}
	\label{fig:model_architecture}
    \vspace{-10pt}
\end{figure*}

\noindent
\textbf{Semantic Image Segmentation.}
While early efforts focused on structural models with handcrafted features~\cite{kae2013augmenting, koltun2011efficient, vineet2012filter, yang2014context}, recent studies employ deep convolutional neural network (CNN) to learning strong representation, which improves segmentation accuracy significantly~\cite{chen2014semantic,liu2015semantic,liu2016deep,long2014fully,zheng2015conditional}.
For instance, Long~\etal \cite{long2014fully} transformed fully-connected layers of CNN into convolutional layers, making accurate per-pixel classification possible using the contemporary CNN architectures that were pre-trained on ImageNet~\cite{deng2009imagenet}.
Chen~\etal~\cite{chen2014semantic}, Zheng~\etal~\cite{zheng2015conditional}, and Liu~\etal~\cite{liu2015semantic,liu2016deep} further showed that back-propagation and inference of Markov Random Field (MRF) can be incorporated into CNN.
Though attaining high accuracy, these models generally have high computational costs, preventing them from deploying in real-time.

Another line of research~\cite{badrinarayanan2015segnet, liu2016learning, paszke2016enet} alleviates this problem by using lightweight network architectures.
For example, SegNet~\cite{badrinarayanan2015segnet} adopted a convolutional encoder-decoder and removed unnecessary layers to reduce the number of parameters.
ENet~\cite{paszke2016enet} utilized a bottleneck module to reduce computation of convolutions.
Although these networks are speeded up, they sacrificed high performances as presented in previous deep models.
This work proposes Deep Layer Cascade (LC), which improves both speed and accuracy of existing deep networks.
It achieves state-of-the-art performances on both Pascal VOC and Cityscape datasets, and runs in real time.

\noindent
\textbf{Deep Learning Cascade.}
Network cascades~\cite{cai2015learning, li2015convolutional, murthy2016deep, toshev2014deeppose, liu2016fashion} have been studied to improve the performance in classification~\cite{murthy2016deep}, detection~\cite{li2015convolutional}, and pose estimation~\cite{toshev2014deeppose}.
For example, Deep Decision Network~\cite{murthy2016deep} improved the image classification performance by dividing easy data from the hard ones.
The hard cases with high confusion will be propagated and handled by the subsequent expert networks.
Li \etal~\cite{li2015convolutional} used CNN cascade for face detection, which rejects false detections quickly in early stages and carefully refines detections in later stages.
DeepPose~\cite{toshev2014deeppose} employed a divide-and-conquer strategy and designed a cascaded deep regression framework for human pose estimation.
Different from previous network cascades that train each network separately,  
LC is jointly optimized to boost the segmentation accuracy.

%% file: approach_v9_xxli_zwliu.tex
%
%

Sec.~\ref{sec:turning_into_LC} takes Inception-ResNet-v2~\cite{szegedy2016inception} as an example to illustrate how one could turn a deep model into LC.
The approach can be easily generalized to the other deep networks.
Sec.~\ref{sec:training} introduces the training algorithm of LC.

\subsection{Turning a Deep Model into LC}
\label{sec:turning_into_LC}

\noindent
\textbf{Network Overview.} 
To illustrate the effectiveness of LC, we choose Inception-ResNet-v2 pre-trained on ImageNet dataset as a strong baseline, denoted as IRNet, which outperforms ResNet-101 by 1.2\% on the Pascal VOC2012 validation set.
Experiments demonstrate that LC is able to achieve 1.7\% improvement on this competitive baseline.

Figure~\ref{fig:model_architecture} (a) visualizes the architecture of IRNet, which has six different components, including `Stem', `IRNet-A/B/C', and `Reduction-A/B'.
Different components have different configurations of layers, such as convolution, pooling, and concatenation layer.
The right column of Fig.~\ref{fig:model_architecture} shows the structures of `Stem' and `IRNet-A/B/C' respectively, including layer types, kernel sizes, and the number of channels (in bracket).
The stride typical equals one unless otherwise stated.
For example, `Stem' employs an RGB image as an input and produces features of 384 channels.
More specifically, the input image is forwarded to three convolutional layers with 3$\times$3 kernels, and then the learned features are split into two streams, which have 3 and 5 convolutional layers respectively.
%
%

%
Similar network structure as IRNet has achieved great success in image recognition \cite{szegedy2016inception}.
However, two important modifications are necessary to adapt it to image segmentation.
Firstly, to increase the resolution of prediction, we remove the pooling layer at the end of IRNet and enlarge the size of feature maps by decreasing the convolutional strides in `Reduction-A/B' (from 2 to 1).
In this case, we expand the size of network outputs (label maps) by 4$\times$.
We also replace convolutions in `IRNet-B/C' by the dilated convolutions similar to \cite{chen2014semantic}.
Secondly, as feature maps with high resolution consume a large amount of GPU memory in the learning process, they limit the size of mini-batch (\eg 8), making the batch normalization (BN) layers \cite{ioffe2015batch} unstable (as which need to estimate sample mean and variance from the data in a mini-batch).
We cope with this issue by simply fixing the values of all parameters in BNs.
This strategy works well in practice.

\noindent
\textbf{From IRNet to LC (IRNet-LC).}
IRNet is turned into LC by dividing its different components as different stages.
The number of stages is three, which is a common setting in previous cascade methods \cite{li2015convolutional, sun2013deep, toshev2014deeppose}.
As shown in Fig.~\ref{fig:model_architecture} (b), components before `Reduction-A' are considered as the first stage, components between `Reduction-A' and `-B' are the second stage, and the remaining layers become the third stage.
In Fig.~\ref{fig:model_architecture} (b), these three stages are distinguished in yellow, green, and blue respectively.
For instance, stage-1 contains one `Stem', five `IRNet-A', and one `Reduction-A'.
In addition, we append two convolutional layers and a softmax loss at the end of each stage.
In this case, the original IRNet with one loss function develops into multiple stages, where each stage has its own loss function.

Now we introduce the information flows for three stages in IRNet-LC.
In the first stage as shown in Fig.~\ref{fig:model_architecture} (b), given a 3$\times$512$\times$512 image $I$, stage-1 predicts a 21$\times$64$\times$64 segmentation label map $L^1$, where each 21$\times$1 column vector, denoted as $L^1_i\in\mathbb{R}^{21\times1}$, indicates the probabilities (confidence scores) of the $i$-th pixel
belonging to 21 object categories in VOC respectively.
We have $\sum_{j=1}^{21} L^1_{ij}=1$, which can be satisfied by using the softmax function.
If the maximum score of the $i$-th pixel, $\ell^1_i=\max(L^1_{i})$ and $\ell^1_i\in\{L^1_{ij}|j=1...21\}$, is larger than a threshold $\rho$ ($\ell^1_i\geq\rho$), we accept its prediction and do not propagate it forward to stage-2.
%
The value of $\rho$ is usually larger than 0.95.
As introduced in Sec.~\ref{sec:intro}, those pixels in stage-1 that fulfil $\ell^1\geq0.95$ occupy nearly 40\% region of an image, containing a lot of easy pixels and a small number of extremely hard pixels that have high confidence to be misclassified.
Removing them from the network significantly reduces computations and improves accuracy, by
enabling deeper layers to focus on foreground objects.

Stage-2 strictly follows the same procedure as above to determine which pixel is forwarded to stage-3.
In other words, LC only introduces one hyper-parameter $\rho$ to IRNet.
In our implementation, the value of $\rho$ is the same for both stage-1 and -2.
Specifically, $\rho$ represents how many easy and extremely hard pixels are rejected (discarded) in each stage.
A larger value of $\rho$ rejects a smaller number of pixels, whilst smaller $\rho$ discards more pixels.
To the extreme, when $\rho=1.0$, no pixels are rejected. IRNet-LC becomes the original IRNet.
When $\rho=0.9$, 52\% and 35\% pixels are discarded in stage-1 and -2 respectively.

However, if $\rho$ becomes smaller, \ie $\rho<0.9$, more `moderate' pixels that locate on the important parts of objects are discarded, hindering the performance of the deep model.
Experiments show that IRNet-LC is robust when $\rho\in[0.9,1.0]$.
For example, when $\rho=0.95$, IRNet-LC obtains nearly realtime of 18 FPS compared to 9 FPS of IRNet, while outperforms it by 0.8\% accuracy on VOC val.
When $\rho=0.985$, IRNet-LC improves IRNet by 1.7\% with a speed of 15 FPS.
%

After propagating an image through all three stages, we directly combine the predicted label maps of these stages as the final prediction, because different stages predict different regions.
For example, as shown in Fig.~\ref{fig:model_architecture} (b), stage-1 trusts the predictions in most of the `background' (pixels with $\ell_i^1\geq\rho$) and propagates the other region forward.
Pixels in this region are marked as `unknown' because $\ell_i^1<\rho$.
In stage-2, `IRNet-B' and `Reduction-B' only compute convolutions with respect to the forwarded region.
It is learned to predict `harder' region, such as `person' and `horse'.
This process is repeated in stage-3.
\subsection{Region Convolution}
\label{subsec:region_conv}

\begin{figure}[t]
    \centering
    \includegraphics[width=0.4\textwidth]{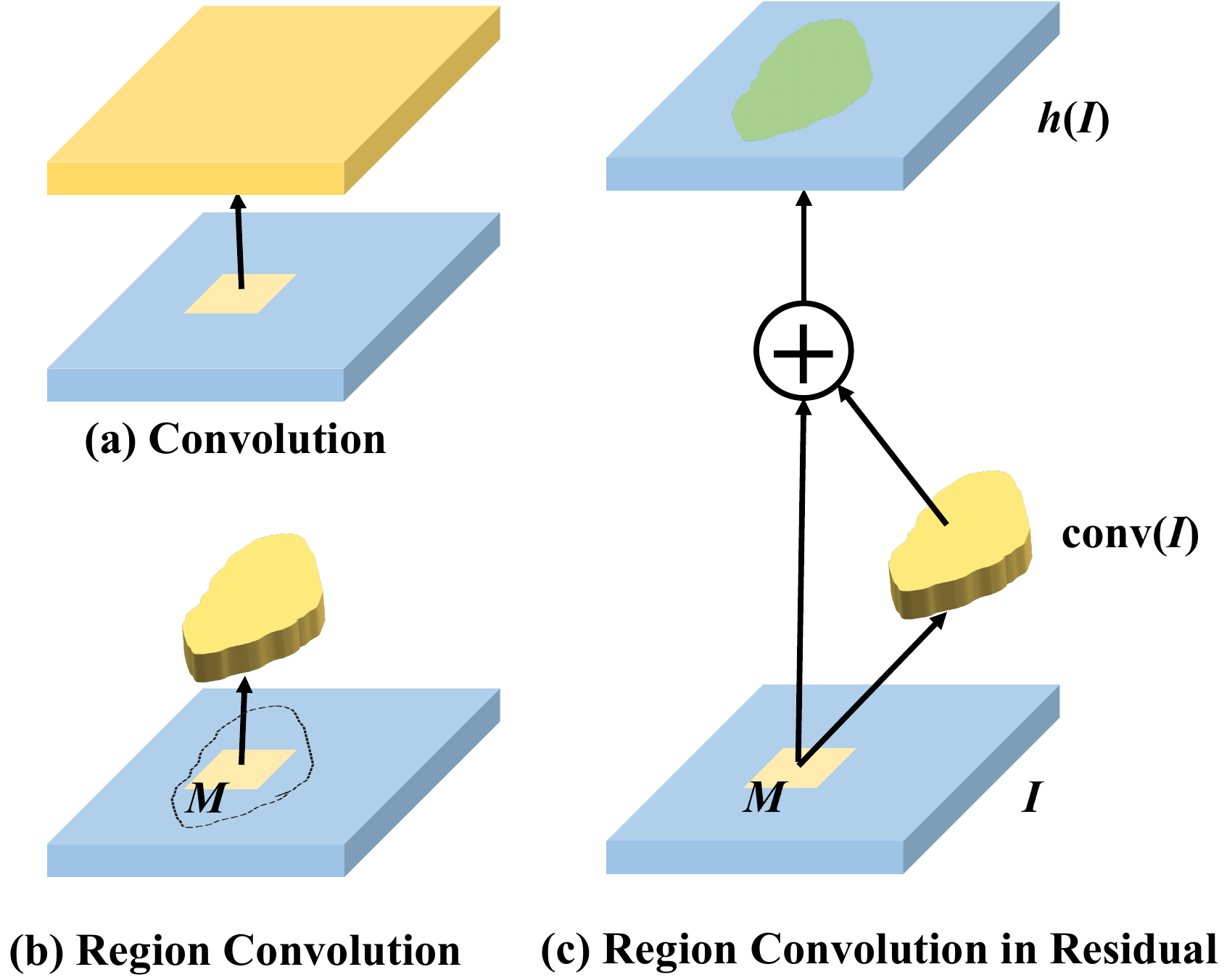}
    \vskip -0.2cm
    \caption{\small{(a) shows the conventional convolution that operates on an entire image. (b) is region convolution (RC) where filters only convolve irregular region of interest denoted as $M$. Values of the other region are set as zeros. (c) illustrates RC in a residual module. \textbf{Best viewed in color.}}}
    \label{fig:region_conv}
    \vspace{-12pt}
\end{figure}

As presented above, stage-2 and -3 only calculate convolutions on those pixels that have been propagated forward.
Fig.~\ref{fig:region_conv}(b) illustrates this region convolution (RC) compared to the traditional convolution in (a), which is applied on an entire feature map.
The filters in RC only convolves a region of interest, denoted as $M$, and ignores the other region, reducing computations a lot. The values of the other region are directly set as zeros.
$M$ can be implemented as a binary mask,
where the pixels inside $M$ equal one, otherwise zero.

Specifically, (c) shows how to apply RC on a residual module, which can be represented as $h(I) = I+ \mathrm{conv}(I)$, where feature $h$ is attained by an identity mapping~\cite{He2015} of $I$ and a convolution over $I$.
We replace the conventional convolution with a RC as introduced above, and the feature $h'(I)$ is the elementwise sum between $I$ and the output of RC.
This is equivalent to learn a masked residual representation, where values inside $M$ are the outputs of RC and those outside $M$ are copied from $I$.
It works well because different stages in LC cope with different non-overlapping regions, and each stage only needs to learn features of regions it concerns.

\subsection{Training IRNet-LC}\label{sec:training}
The parameters of IRNet are initialized by pre-training in ImageNet.
Since IRNet-LC has additional convolutional layers stacked before each loss function, their parameters are initialized by sampling from a normal distribution.
Given a set of images and their per-pixel label maps, IRNet-LC is learned in two steps, where the first one aims at initial training and the second one employs cascade training.


\noindent
\textbf{Initial Training.}
This step is similar to deeply supervised network (DSN)~\cite{lee2015deeply}, which has multiple identical loss functions in different layers of the network.
Its objective is to adapt IRNet pre-trained by classifying one thousand image categories in ImageNet to the task of image segmentation.
It learns discriminative and robust features.
%
In IRNet-LC, every stage is trained to minimize a pixel-wise softmax loss function, measuring the discrepancies between the predicted label map and the ground truth label map of the entire image.
These loss functions are jointly optimized by using back-propagation (BP) and stochastic gradient descent (SGD).

%


\noindent
\textbf{Cascade Training.}
Once we finish the initial training, we fine-tune each stage of IRNet-LC by leveraging the cascade strategy of $\rho$ as introduced in Sec.~\ref{sec:turning_into_LC}.
Similar to the previous step, all stages are trained jointly, but different stages minimize their pixel-wise softmax losses with respect to different regions.
More specific, the gradients in BP are only propagated to the region of interest in each stage, which is able to learn discriminative features corresponding to regions (pixels) in a specific difficulty-level.
Intuitively, the current stage is fine-tuned on pixels that have low confidences in the previous stage,
enabling `harder' pixels to be captured by deeper layers to improve segmentation accuracy and reduce computation.

\noindent
\textbf{Training Details.}
We fix a mini-batch size of $12$ images, momentum $0.9$ and weight decay of $0.0005$ for both two steps.
In the initial training, we set the initial learning rate to be $10^{-4}$ and drop it by a factor of 10 after every 10 epochs.
In the cascade training, we also set the initial learning rate to be $10^{-4}$ and drop it by a factor of 10 after every 15 epochs.

\subsection{Relations with Previous Models}
\label{subsec:relation_models}

The relationships and differences between LC and MC have been discussed in Sec.~\ref{sec:intro}.
LC also relates to deeply supervised nets (DSN)~\cite{lee2015deeply} and dropout~\cite{srivastava2014dropout}.


\noindent
\textbf{DSN.}
Similar to DSN, LC adds supervision to each stage.
However, to enable adaptive processing of hard/easy regions, LC employs different supervisions for different stages.
In contrast, the supervision used in each stage of DSN are kept the same.
Specifically, the stage-wise supervision in LC is determined by the estimated difficulty of each pixel.
In this way, each stage in LC is able to focus on regions with a similar difficulty level.
%

\noindent
\textbf{Dropout.}
LC connects to dropout in the sense that both methods discard some regions in the feature maps, but they are essentially different.
LC drops those pixels with high confidences and only propagates difficult pixels forward to succeeding stages.
The easy and ambiguous regions are perpetually dropped in upper layers so as to reduce computations and the deeper layers focus more on `hard' regions such as foreground objects.
Dropout randomly zeros out pixels in each layer independently.
It prevents over-fitting but slightly increases computations.
In the experiment, LC is compared with dropout to identify that the performance gain mainly comes from the proposed cascade strategy.





%% file: experiments_v8_cavan_zwliu.tex
\begin{figure}
    \centering
    \includegraphics[width=0.48\textwidth]{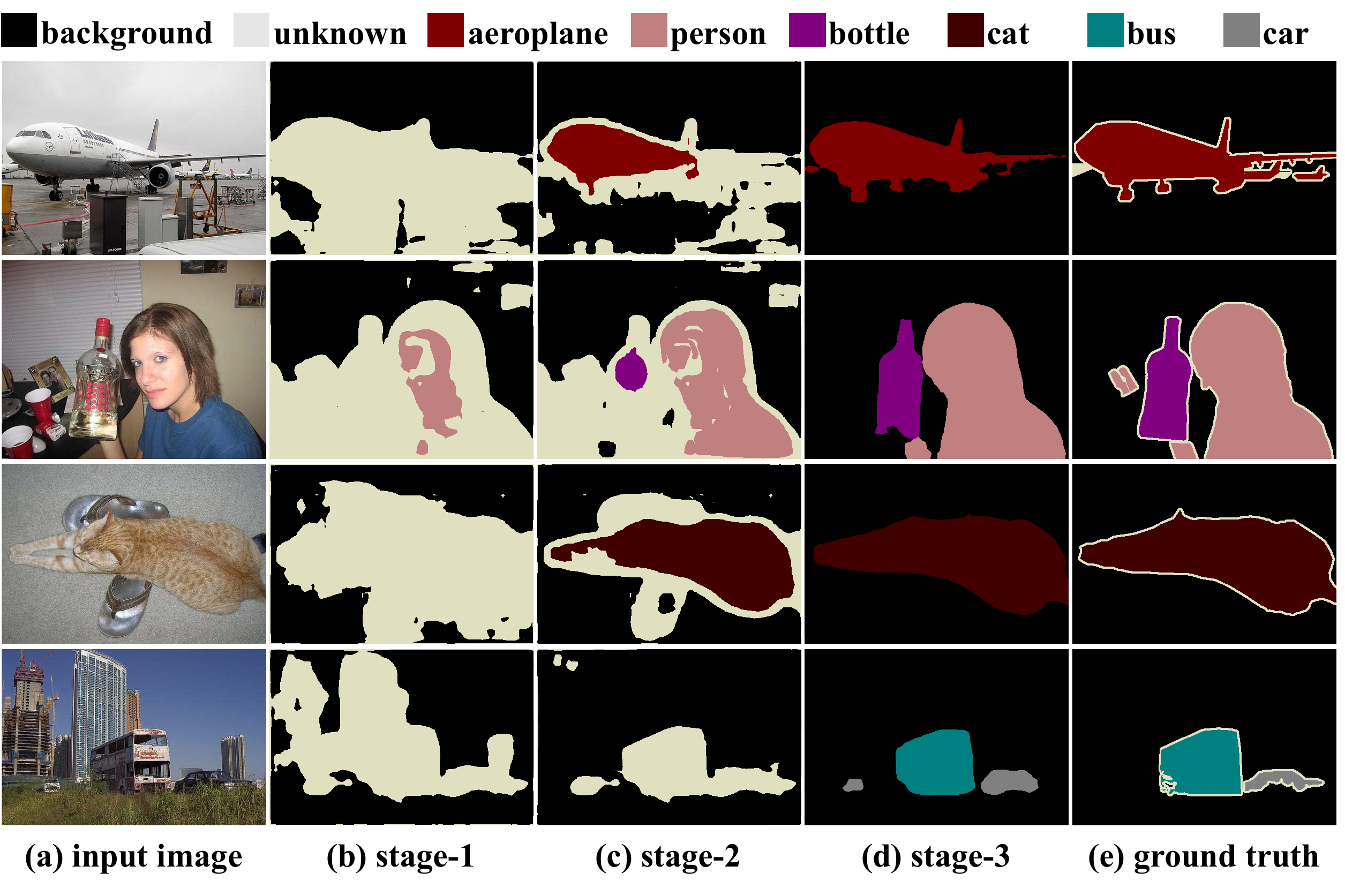}
    \caption{\small{Visualization of different stages' outputs in VOC12 dataset. \textbf{Best viewed in color.}}}
    \label{fig:stage_results}
    \vspace{-12pt}
\end{figure}

\noindent
\textbf{Settings.} 
We evaluate our method on the PASCAL VOC 2012 (VOC12)~\cite{everingham2010pascal} and Cityscapes~\cite{Cordts2016Cityscapes} datasets.
VOC12 dataset is a generic object segmentation benchmark with 21 classes.
Following previous works, we also use the extra annotations provided by~\cite{hariharan2011semantic}, which contains $10,582$ images for training, $1,449$ images for validation, and $1,456$ images for testing. 
Cityscapes dataset, on the other hand, focuses on street scenes segmentation and contains 19 categories.
%
%
In our experiments, we only employ images with fine pixel-level annotations. There are 2975 training, 500 validation and 1525 testing images. This is consistent with existing studies~\cite{lin2015efficient, CP2016Deeplab}.
We adopt mean intersection over union (mIoU) to evaluate the performance of different methods. 

\subsection{Ablation Study}
\label{sec:effectiveness}


In this section, we investigate the effects of adjusting probability threshold in LC and demonstrate the merits of LC by comparing it to other counterparts.
All performance are reported on the \emph{validation set} of VOC12.

\noindent
\textbf{Probability Thresholds.}
In each stage of LC, we employ a pixel-wise probability from softmax layer to represent the confidence of prediction.
%
By choosing appropriate probability threshold $\rho$, LC can separate easy regions, moderate regions and extremely hard regions for adaptive processing.
As discussed in Sec.~\ref{sec:turning_into_LC}, $\rho$ controls how many easy and extremely hard pixels are discarded in each stage.

Table \ref{tab:ablation_threshold} lists the processed pixel percentage in stage-1 \& -2 and the overall performance as $\rho$ varies.
If $\rho = 1$, LC will degenerate to DSN, which is slightly better than fully convolutional IRNet.
When $\rho$ decreases, more easy regions are classified in early stages while hard regions are progressively handled by later stages.
It can be understood as hard negative mining~\cite{girshick2014rich, shrivastava2016training} which improves the performance.
On the other hand, if the value of $\rho$ is too small, the algorithm might become too optimistic, \ie many hard regions are processed in early stages and early decisions are made.
The performance will be harmed by overly early decisions when hard regions do not receive sufficient inference using deeper layers.
As shown in Table \ref{tab:ablation_threshold}, when $\rho = 0.985$, \ie, LC processes around 52\% regions in early stages and achieves the best performance. This value is used in all the following experiments.
In practice, the value of $\rho$ can be chosen empirically using a validation set.


\begin{table}
    \small
\caption{Ablation study on probability thresholds $\rho$.}
    \centering
    \begin{tabular}{@{}c@{ \,}|@{\,}c@{ \,}|@{\,}c@{ \,}|@{\,}c@{ \,}|@{\,}c@{ \,}|@{\,}c@{ \,}|@{\,}c@{ \,}|@{\,}c@{ \,}|@{\,}c@{ \,}}
        \hline
        $\rho$ & 1&0.995&0.985&0.970&0.950&0.930&0.900&0.800\\
        \hline\hline
        stage-1 (\%) & 0& 15&  23& 30& 35& 35& 44& 56\\
        stage-2 (\%) & 0& 14&  29& 31& 30& 41& 31& 29\\
        \hline
        mIoU (\%) &72.70&73.56& \textbf{73.91} & 73.63& 73.03&72.53&71.20&66.95\\
        \hline
    \end{tabular}
    \label{tab:ablation_threshold}
\end{table}

\begin{table}
    \small
    \caption{Comparisons with related methods.}
    \centering
    \begin{tabular}{l|c}
        \hline
        &mIoU(\%)\\
        \hline\hline
        IRNet~\cite{szegedy2016inception} & 72.22\\ 
        DSN~\cite{lee2015deeply} &72.70\\  
        DSN~\cite{lee2015deeply} + Dropout~\cite{srivastava2014dropout} & 72.63\\  
        Model Cascade (MC) & 44.20\\ 
        \hline\hline
        Layer Cascade (LC) & \textbf{73.91} \\
        \hline
    \end{tabular}
    \label{tab:related_methods}
\end{table}

\vspace{0.1cm}
\noindent
\textbf{Effectiveness of Layer Cascade.}
To show the merits of LC, we compare it to some important counterparts as discussed in Sec.~\ref{subsec:relation_models}, including:
\begin{itemize}
\small{
    \item \textbf{IRNet~\cite{szegedy2016inception}:}
    We use the model describe in Sec.~\ref{sec:turning_into_LC} as baseline. 
    To conduct a fair comparison, all the following methods are based on this backbone network.
    \item \textbf{DSN~\cite{lee2015deeply}:}
    By setting $\rho$ = 1, we make LC degenerate to a DSN, where each stage process all regions and has full supervision as the final target.
    \item \textbf{DSN~\cite{lee2015deeply} + Dropout~\cite{srivastava2014dropout}:}
    To distinguish our method from dropout, LC is compared against DSN equipped with random label dropout in each stage.
    We keep the dropout ratio identical as that in LC.
    \item \textbf{Model Cascade:}
    MC has a similar network architecture to LC, but with different training strategy as discussed in Sec.~\ref{sec:intro}. 
    Specifically, MC divides the IRNet into three stages, and each stage is trained separately.
    When we train a certain stage, we fix the parameters of all previous stages.
    The same threshold as in LC is employed here, \ie, $\rho = 0.985$.  
}
\end{itemize}

The results are summarized in Table \ref{tab:related_methods}. We have three observations here.
Firstly, the improvement from deep supervision (DSN) is relatively limited, which only leads to $0.48$ mIoU gain in comparison to the baseline IRNet.
Since pre-training on ImageNet has been a common practice in semantic segmentation~\cite{long2014fully}, which effectively prevents gradients exploding or vanishing, it renders the advantages of deeply supervision marginal.
%
Secondly, random label dropout does not bring significant effect to the result.
The result is expected because the dropout technique is designed to alleviate the hazard of overfitting given small training data size.
However, semantic segmentation is a per-pixel labeling task and we have abundant training data to support the learning task.
Thirdly, Model Cascade (MC) performs even worse than the baseline IRNet.
It is because MC divides the IRNet into several independent sub-models. But each sub-model is shallow and therefore weaken the overall modeling capacity.  
On the contrary, LC has the appealing properties of cascading and also keeping the \textit{intrinsic depth} for the whole model. 
The capability of maintaining the model depth adaptively for hard regions makes our approach outstanding in the comparison.

\begin{figure}
    \centering
    \includegraphics[width=0.5\textwidth]{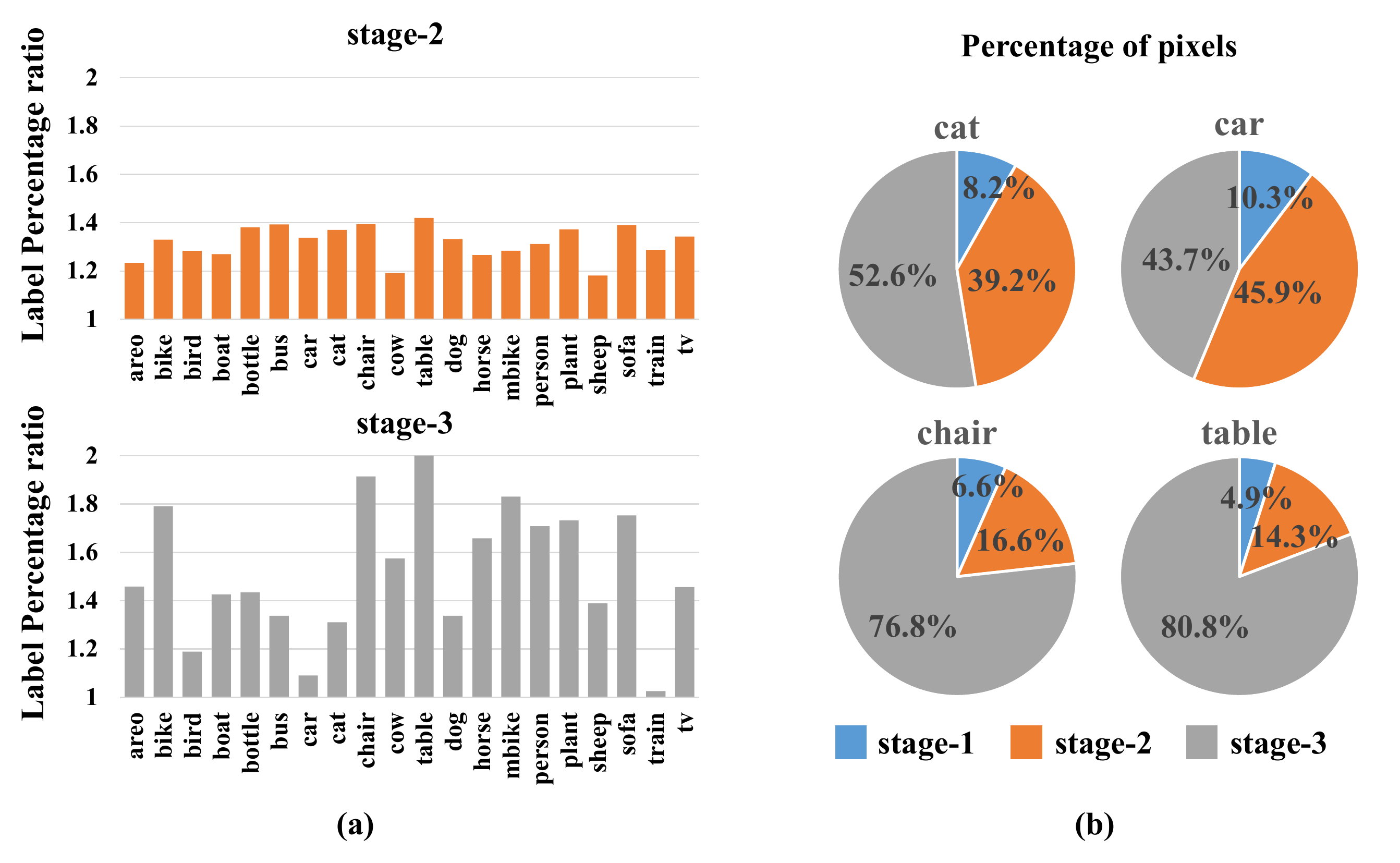}
    \vskip -0.2cm
    \caption{\small{(a) is the change of label distribution in stage-2 and -3. (b) shows the percentage of pixels that are classified in different stages.}}
    \label{fig:data_balance}
    \vspace{-12pt}
\end{figure}

\subsection{Stage-wise Analysis}
\label{subsec:stage_analysis}

In this section, we demonstrate how LC enables adaptive processing for different classes and visualize the regions handled by different regions. 

\begin{figure*}
    \centering
    \includegraphics[width=1.0\textwidth]{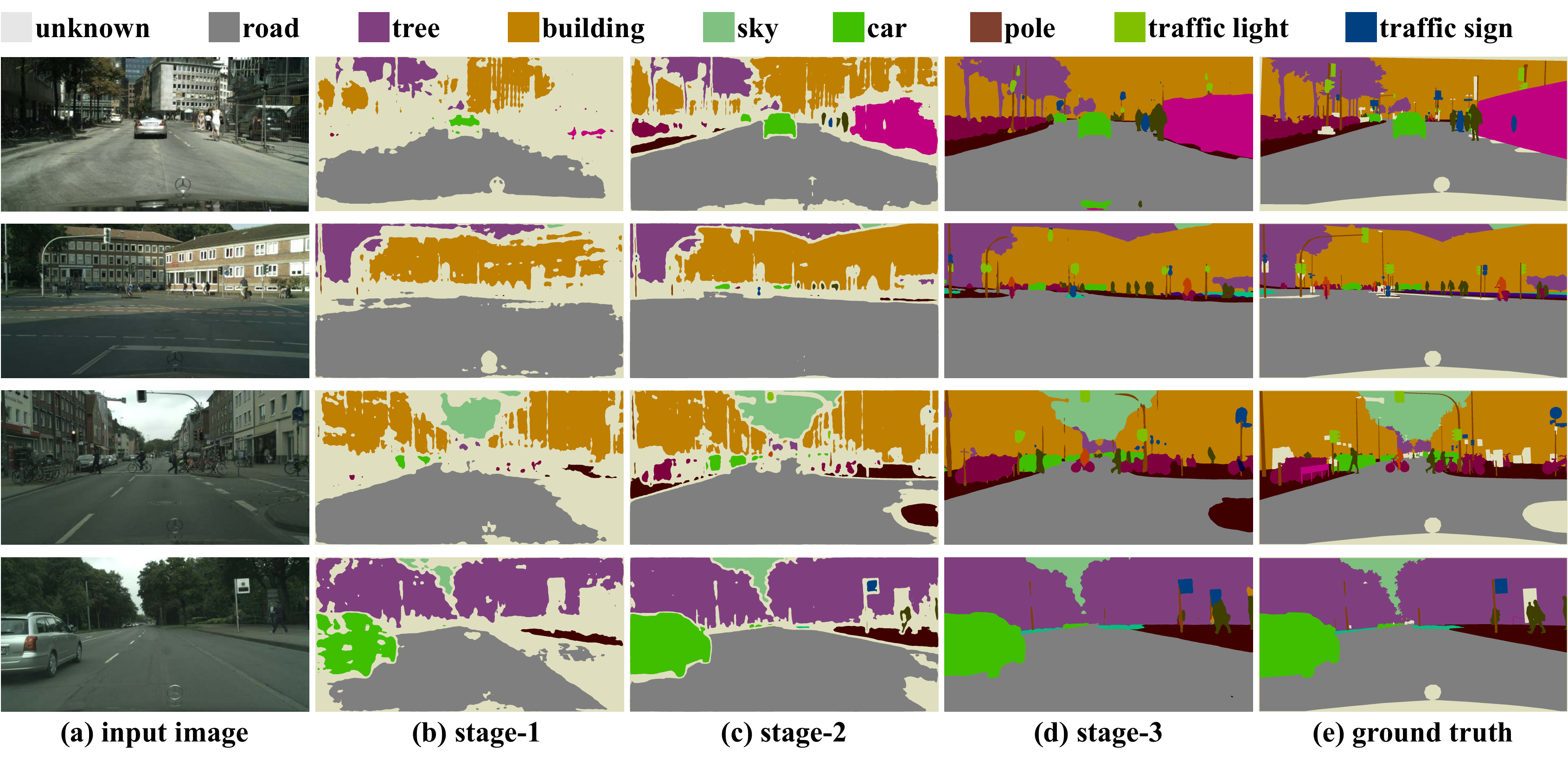}
    \vskip -0.2cm
    \caption{\small{Visualization of different stages' outputs in Cityscapes dataset. \textbf{Best viewed in color.}}}
    \label{fig:stage_results_cityscapes}
\end{figure*}

\noindent
\textbf{Stage-wise Label Distribution.}
First, we provide a label distribution analysis across different stages. 
Here we take the $20$ classes (excluding ``background'') in VOC12 as an example.
Fig.~\ref{fig:data_balance} (a) shows how the number of pixels changes with respect to each class in stage-2 and -3. For example, the upper histogram shows a ratio for each class, obtained by dividing its number of pixels in stage-2 by those in stage-1. Ratios larger than one indicates stage-2 focus more on the corresponding classes than stage-1 does. 
We find that all ratios have increased and belong to the range of $1$ to $1.4$.
It is because stage-1 already handles the easy regions (\ie ``background'') and leaves the hard regions (\ie ``foreground'') to stage-2.
Ratios of stage-3 can be obtained similarly in the bottom histogram. When comparing stage-3 to -2, we can see that stage-3 further focus on harder classes (\eg ``bicycle'', ``chair'' and ``dining table'').
LC learns to process samples in a ``difficulty-aware'' manner.
We also conduct a per-class analysis as illustrated in Fig.~\ref{fig:data_balance} (b).
Harder classes like ``chair'' and ``table'' have more pixels handled by deeper layers (stage-3).

\noindent
\textbf{Stage-wise Visualization.}
Here we visualize the output label maps of different stages for both VOC12 and Cityscapes, as shown in Fig.~\ref{fig:stage_results} and ~\ref{fig:stage_results_cityscapes}.
The uncertain regions in different stages are also marked out.
In VOC12, the easy regions like ``background'' and ``human faces'' are first labeled by stage-1 in LC.
The remaining foreground and boundary regions are then progressively labeled by stage-2 and stage-3 in LC. 
Similarly, in Cityscapes, the easy regions like ``road'' and ``building'' are first labeled by stage-1.
Other small objects and fine details like ``pole'' and ``pedestrian'' are handled by stage-2 and -3.

\subsection{Performance and Speed Analysis}

%

\noindent
\textbf{Comparisons with DeepLab and SegNet.}
To highlight the trade-off between performance and speed, we compare the proposed LC model with two representative state-of-the-art methods, DeepLab-v2~\cite{CP2016Deeplab} and SegNet~\cite{badrinarayanan2015segnet}.
The performance are reported on VOC12 and summarized in Table \ref{tab:region_conv}. The runtime speed is measured on a single Titan X GPU.
To ensure a fair comparison, we evaluate DeepLab-v2 and SegNet without any pre- and post-processing, \eg, training with extra data, multi-scale fusion, or smoothing with conditional random fields (CRF).

DeepLab-v2 achieves an acceptable mIoU of $70.42$.
Nonetheless, it uses an ultra-deep ResNet-101 model as the backbone network, its speed of inference is thus slow ($7.1$ FPS).
On the contrary, SegNet is faster due to a smaller model size, however, its accuracy is greatly compromised. In particular, it increases its speed to $14.6$ FPS through sacrificing of over $10$ mIoU.
%
The proposed LC alleviates the need of trading-off speed with a large drop in performance. The cascaded end-to-end trainable framework with region convolution allows it to achieve the best performance ($73.91$ mIoU) with an acceptable speed ($14.7$ FPS).

\begin{table}[t]
    \small
\caption{A comparison of performance and speed of Layer Cascade (LC) against existing methods.}
    \centering
    \begin{tabular}{l|c|c|c}
        \hline
        &mIoU&ms&FPS\\
        \hline\hline
        DeepLab-v2~\cite{CP2016Deeplab}&70.42&140.0&7.1\\
        SegNet~\cite{badrinarayanan2015segnet}&59.90&69.0&14.6\\
        \hline\hline    
        LC &\textbf{73.91}&65.1&14.7\\
        LC (fast)&66.95&42.5&\textbf{23.6}\\
        \hline
    \end{tabular}
    \label{tab:region_conv}
\end{table}

\begin{figure}[t]
    \centering
    \includegraphics[width=0.5\textwidth]{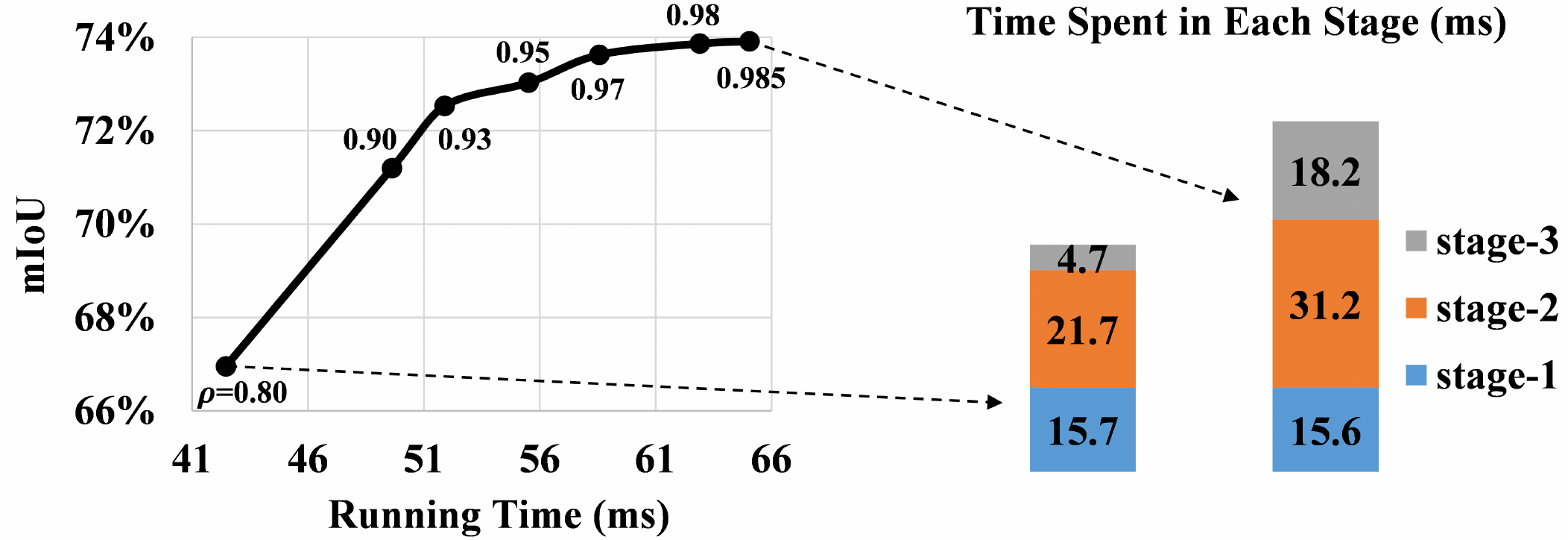}
    \vskip -0.2cm
    \caption{\small{(a) shows the performance and speed trade-off in Layer Cascade (LC) by adjusting $\rho$. (b) is the time spent in each stage.}}
    \label{fig:trade_off}
\end{figure}

\noindent
\textbf{Further Performance and Speed Trade-off.}
It is worth pointing out that the runtime of LC can be further reduced by decreasing $\rho$ to allow more regions to be handled by early stages.
The performance and speed trade-off is depicted in Fig.~\ref{fig:trade_off} (a) with the corresponding $\rho$ values. It is observed that decreasing $\rho$ slightly affects the accuracy, but it greatly reduces the computation time.
Notably, when LC attains real-time inference at $23.6$ FPS, it still exhibits competitive mIoU of $66.95$, in comparison to mIoU of $70.42$ yielded by at $7.1$ FPS. 
We also include the per-stage runtime in Fig.~\ref{fig:trade_off} (b).
The increasing computation for higher performance mainly comes from later stages.

\begin{table*}[t]
    \caption{Per-class results on VOC12 \textit{test set}. Approaches pre-trained on COCO \cite{lin2014microsoft} are marked with $^\dagger$.}
    \scriptsize
    \centering
    \begin{tabular}{@{}l@{\,}|p{9pt}p{9pt}p{9pt}p{9pt}p{9pt}p{9pt}p{9pt}p{9pt}p{9pt}p{9pt}p{9pt}p{9pt}p{9pt}p{9pt}p{9pt}p{9pt}p{9pt}p{9pt}p{9pt}p{9pt}|p{12pt}}
        \hline
        & areo & bike & bird & boat & bottle & bus & car & cat & chair & cow & table & dog & horse & mbike & person & plant & sheep & sofa & train & tv & mIoU \\
        \hline\hline
        FCN \cite{long2014fully} & 76.8 & 34.2 & 68.9 & 49.4 & 60.3 & 75.3 & 74.7 & 77.6 & 21.4 & 62.5 & 46.8 & 71.8 & 63.9 & 76.5 & 73.9 & 45.2 & 72.4 & 37.4 & 70.9 & 55.1 & 62.2 \\
        DeepLab \cite{chen2014semantic} & 84.4 & 54.5 & 81.5 & 63.6 & 65.9 & 85.1 & 79.1 & 83.4 & 30.7 & 74.1 & 59.8 & 79.0 & 76.1 & 83.2 & 80.8 & 59.7 & 82.2 & 50.4 & 73.1 & 63.7 & 71.6 \\
        RNN \cite{zheng2015conditional} & 87.5 & 39.0 & 79.7 & 64.2 & 68.3 & 87.6 & 80.8 & 84.4 & 30.4 & 78.2 & 60.4 & 80.5 & 77.8 & 83.1 & 80.6 & 59.5 & 82.8 & 47.8 & 78.3 & 67.1 & 72.0 \\
        Adelaide \cite{wu2016high} &91.9&48.1&93.4&\textbf{69.3}&75.5&94.2&87.5&92.8&36.7&86.9&65.2&89.1&90.2&86.5&87.2&64.6&\textbf{90.1}&59.7&85.5&72.7&79.1\\
        \hline
        RNN$^\dagger$ \cite{zheng2015conditional} & 90.4 & 55.3 & 88.7 & 68.4 & 69.8 & 88.3 & 82.4 & 85.1 & 32.6 & 78.5 & 64.4 & 79.6 & 81.9 & 86.4 & 81.8 & 58.6 & 82.4 & 53.5 & 77.4 & 70.1 & 74.7 \\
        BoxSup$^\dagger$ \cite{dai2015boxsup} & 89.8 & 38.0 & 89.2 & 68.9 & 68.0 & 89.6 & 83.0 & 87.7 & 34.4 & 83.6 & 67.1 & 81.5 & 83.7 & 85.2 & 83.5 & 58.6 & 84.9 & 55.8 & 81.2 & 70.7 & 75.2 \\
        DPN$^\dagger$ \cite{liu2015semantic} & 89.0 & 61.6 & 87.7 & 66.8 & 74.7 & 91.2 & 84.3 & 87.6 & 36.5 & 86.3 & 66.1 & 84.4 & 87.8 & 85.6 & 85.4 & 63.6 & 87.3 & 61.3 & 79.4 & 66.4 & 77.5 \\
        DeepLab-v2$^\dagger$ \cite{CP2016Deeplab} &92.6&60.4&91.6&63.4&76.3&95.0&88.4&92.6&32.7&88.5&67.6&89.6&\textbf{92.1}&87.0&87.4&63.3&88.3&60.0&\textbf{86.8}&74.5&79.7\\
        \hline\hline
        LC & \textbf{94.1} & 63.0 & 91.2 & 67.9 & 79.5 & 93.4 & \textbf{90.0} & \textbf{93.8} & 37.4 & 83.7 & 65.9 & \textbf{90.7} & 86.1 & 88.8 & 87.5 & 68.5 & 86.9 & 64.3 & 85.6 & 72.2 & \textbf{80.3} \\
        LC$^\dagger$ & 85.5 & \textbf{66.7} & \textbf{94.5} & 67.2 & \textbf{84.0} & \textbf{96.1} & 89.8 & 93.5 & \textbf{47.2} & \textbf{90.4} & \textbf{71.5} & 88.9 & 91.7 & \textbf{89.2} & \textbf{89.1} & \textbf{70.4} & 89.4 & \textbf{70.7} & 84.2 & \textbf{79.6} & \textbf{82.7} \\
        \hline
    \end{tabular}
    \label{tab:perclass}
\end{table*}

\begin{table*}[t]
\caption{ Per-class results on Cityscapes \textit{test set}. ``sub'' denotes whether the method used subsampling images for training.}
    \scriptsize
    \centering
    \begin{tabular}{@{}l@{\,}|p{8pt}|p{8pt}p{8pt}p{8pt}p{8pt}p{8pt}p{8pt}p{8pt}p{8pt}p{8pt}p{8pt}p{8pt}p{8pt}p{8pt}p{8pt}p{8pt}p{8pt}p{8pt}p{8pt}p{8pt}|p{12pt}}
        \hline
        &sub&road&swalk&build.&wall&fence&pole&tlight&sign&veg.&terrain&sky&person&rider&car&truck&bus&train&mbike&bike&mIoU\\
        \hline\hline
        RNN  \cite{zheng2015conditional}  &2&96.3&73.9&88.2&47.6&41.3&35.2&49.5&59.7&90.6&66.1&93.5&70.4&34.7&90.1&39.2&57.5& 55.4 &43.9&54.6&62.5\\
        DeepLab \cite{chen2014semantic} &2&97.3&77.7&87.7&43.6&40.5&29.7&44.5&55.4&89.4&67.0&92.7&71.2&49.4&91.4&48.7&56.7&49.1&47.9&58.6&63.1\\
        FCN \cite{long2014fully} &no&97.4&78.4&89.2&34.9&44.2&47.4&60.1&65&91.4&69.3&93.9&77.1&51.4&92.6&35.3&48.6&46.5&51.6&66.8&65.3\\
        DPN \cite{liu2015semantic} &no&97.5&78.5&89.5&40.4&45.9&51.1&56.8&65.3&91.5&69.4&\textbf{94.5}&77.5&54.2&92.5&44.5&53.4&49.9&52.1&64.8&66.8\\
        Dilation10 \cite{yu2015multi} &no& 97.6 & 79.2 & 89.9 &37.3&47.6&53.2&58.6&65.2& 91.8 &69.4&93.7& 78.9 & 55 &93.3&45.5&53.4&47.7&52.2&66&67.1\\
        DeepLab-v2 \cite{CP2016Deeplab} & no & 97.8 & 81.3 & 90.3 & 48.7 & 47.3 & 49.5 & 57.8 & 67.2 & 91.8 & 69.4 & 94.1 & 79.8 & 59.8 & 93.7 & 56.5 & 67.4 & \textbf{57.4} & 57.6 & 68.8 & 70.4 \\
        Adelaide \cite{lin2015efficient} &no&\textbf{98.0} & 82.6 & 90.6 & 44.0 & 50.7 & 51.1&\textbf{65.0}&71.7&\textbf{92.0}&\textbf{72.0}&94.1&\textbf{81.5}&\textbf{61.1}&\textbf{94.3}&61.1&65.1&53.8&\textbf{61.6}&70.6&\textbf{71.6}\\
        \hline\hline
        LC&no& 97.9 & \textbf{83.1} & \textbf{91.6} & \textbf{53.7} & \textbf{57.4} & \textbf{58.4} & 62.0 & \textbf{73.3} & 91.9&61.3&93.8&78.8&53.1& 93.4 & \textbf{62.2}& \textbf{76.9} & 53.5 & 57.0 & \textbf{74.7} & 71.1 \\
        \hline
    \end{tabular}
    \label{tab:Cityscapes_perclass}
\end{table*}

\subsection{Benchmark}

In this section, we show that LC can achieve state-of-the-art performance on standard benchmarks like VOC12~\cite{everingham2010pascal} and Cityscapes~\cite{Cordts2016Cityscapes} datasets. 
Following~\cite{CP2016Deeplab}, atrous spatial pyramid pooling~\cite{CP2016Deeplab}, three-scale testing and dense CRF~\cite{koltun2011efficient} are employed.

\noindent
\textbf{VOC12.}
Table~\ref{tab:perclass} lists the per-class and overall mean IoU on VOC12 \textit{test set}.
The approaches pre-trained on COCO \cite{lin2014microsoft} are marked with $^\dagger$.
LC achieves a mIoU of $80.3$ 
and further improves the mIoU to $82.7$ 
with pre-training on COCO, which is the best-performing method on VOC12 benchmark.
%
%
By inspecting closer, we observe that LC wins $16$ out of $20$ foreground classes. For other $4$ classes, LC also achieves competitive performance. Large gain is observed in some particular classes such as ``bike'', ``chair'', ``plant'', and ``sofa''. Based on our statistics in Fig.~\ref{fig:data_balance}, we found that these few classes, in general, require a deeper stage to make decisions on hard regions. 
%
%


%
%
%
%

\noindent
\textbf{Cityscapes.}
Next, we evaluate LC on Cityscapes benchmark, with results summarized in Table \ref{tab:Cityscapes_perclass}.
``sub'' denotes whether the method used subsampling images for training.
LC also achieves promising performance with a mIoU of $71.1$, which shows its great generalization ability to diverse objects and scenes.
Lin \etal~\cite{lin2015efficient}'s performance is slightly better than ours, however, LC still wins on 9 out of 19 classes. It is noticed that \cite{lin2015efficient} used a deeper backbone-network and explored richer contextual information. We believe that further performance gain can be achieved if LC is incorporated with these techniques.
LC gains outstanding performance on the classes that are `traditionally regarded' as hard classes, \eg, ``fence'', ``pole'', ``sign'', ``truck'', ``bus'' and ``bike'', which usually exhibit flexible shapes and fine-grained details.
%
%
The results suggest that the end-to-end cascading mechanism in LC is meaningful, especially in alleviating the burden of deeper layers on analyzing easy regions but focusing themselves on hard regions adaptively.
%



\subsection{More Comparisons between IRNet-LC and state-of-the-art Methods}

\begin{table*}[t]
  \caption{Comparisons with state-of-the-art methods on VOC12 \textit{test set}. `-' indicates the corresponding information was not disclosed in the previous papers.}
  \centering
  \begin{tabular}{l|c|c|c|c|c|c|c}
      \hline
      &backbone network&\# params&COCO&multi-scale&MRF/CRF&FPS&mIoU\\
      \hline\hline
      CRF-RNN \cite{zheng2015conditional}&VGG \cite{simonyan2014very}&134.4M&yes&-&yes&-&74.7\\
      DPN \cite{liu2015semantic}&VGG \cite{simonyan2014very} &134.4M&yes&yes&yes&-&77.5\\
      DeepLab-v2 \cite{CP2016Deeplab}&ResNet-101 \cite{He2015}&44.5M&yes&yes&yes&0.9&79.7\\
      \hline\hline
      IRNet-LC&IRNet \cite{szegedy2016inception}&35.5M&no&no&no&\textbf{14.3}&78.2\\
      IRNet-LC&IRNet \cite{szegedy2016inception}&35.5M&no&yes&no&7.7&79.5\\
      IRNet-LC&IRNet \cite{szegedy2016inception}&35.5M&no&yes&yes&1.0&\textbf{80.3}\\
      \hline
  \end{tabular}
  \label{tab:comparisons}
\end{table*}

In Table~\ref{tab:comparisons}, we compare the settings of different best-performing methods on VOC12 \cite{everingham2010pascal} test set, including CRF-RNN \cite{zheng2015conditional}, DPN \cite{liu2015semantic} and DeepLab-v2 \cite{CP2016Deeplab}.
These methods are summarized in terms of `backbone network', `number of parameters', `pre-trained using MS COCO', `multi-scale training/test', `MRF/CRF', `frame per second (FPS)', and `mIOU'.
Note that `-' indicates the corresponding information was not disclosed in previous paper.

IRNet-LC uses Inception-ResNet-v2(IRNet) \cite{szegedy2016inception} as backbone network, which is smaller than ResNet-101 (35.5M \vs 44.5M).
Following DeepLab-v2~\cite{CP2016Deeplab}, atrous spatial pyramid pooling is employed in IRNet-LC.
%
%
%
As shown in Table~\ref{tab:comparisons}, IRNet-LC achieves the best performance even without pre-training on MS COCO \cite{lin2014microsoft}, demonstrating the effectiveness of the Layer Cascade framework. 
When all components of pre- and post-processing such as `COCO', `multiscale', and `CRF' are removed, IRNet-LC still obtains comparable performance with respect to DeepLab-v2 (78.2\% \vs 79.7\%), but significantly outperforms DeepLab-v2 in terms of FPS (14.3 fps \vs 0.9 fps).
In other words, IRNet-LC improves FPS of DeepLab-v2 by 15 times with merely 1.5\% decrease in accuracy.
Note that IRNet-LC outperforms state-of-the-art systems like CRF-RNN and DPN by 3.5\% and 0.7\% respectively, without employing any pre- and post-processing steps.
%

%

\begin{figure*}
	\centering
	\includegraphics[width=1.0\textwidth]{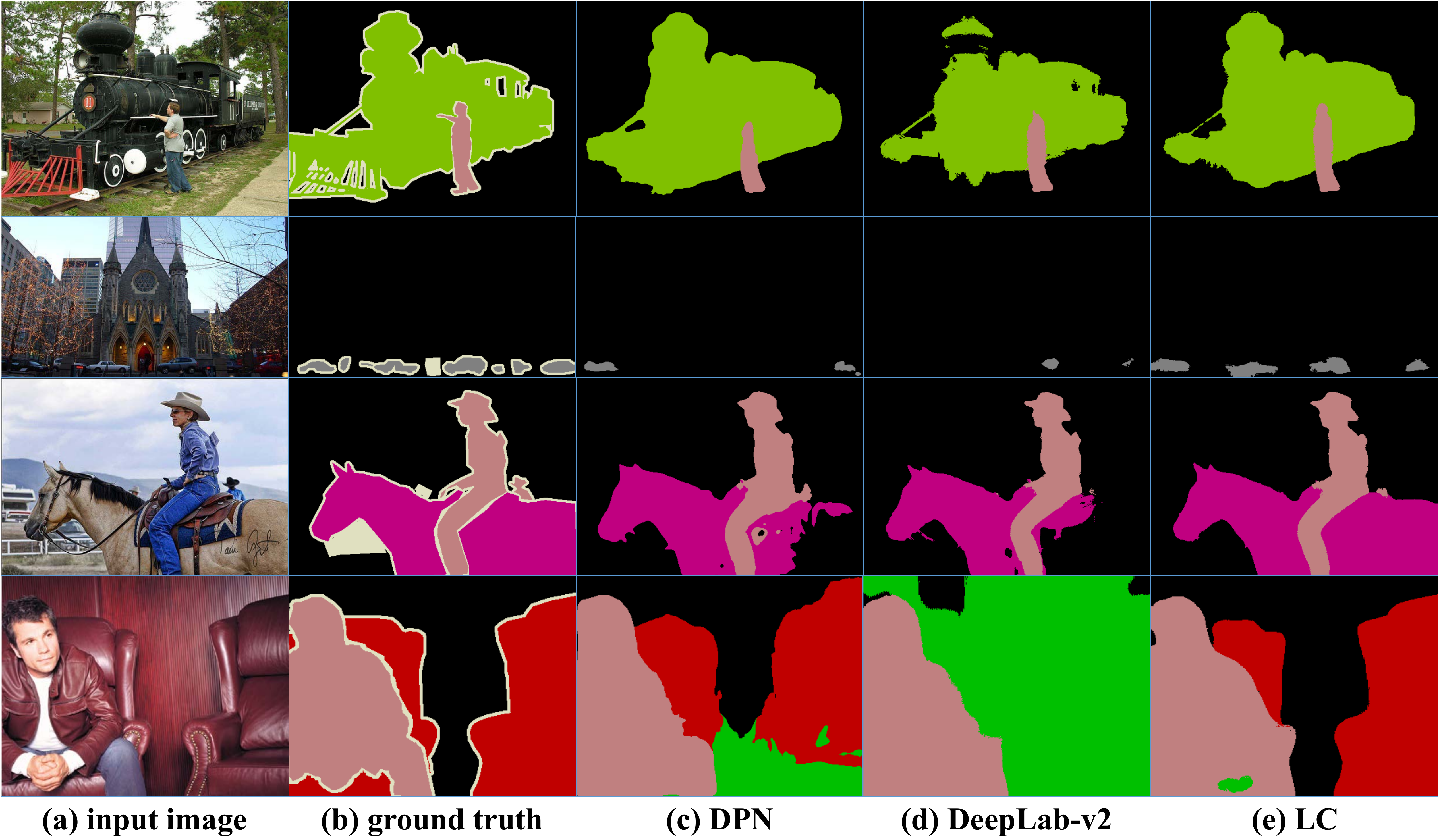}
	\caption{Visual quality comparison of different methods: (a) input image (b) ground truth (c) DPN \cite{liu2015semantic} (d) DeepLab-v2 \cite{CP2016Deeplab} and (e) LC.}
	\label{fig:visual}
\end{figure*}

\begin{figure*}
	\centering
	\includegraphics[width=1.0\textwidth]{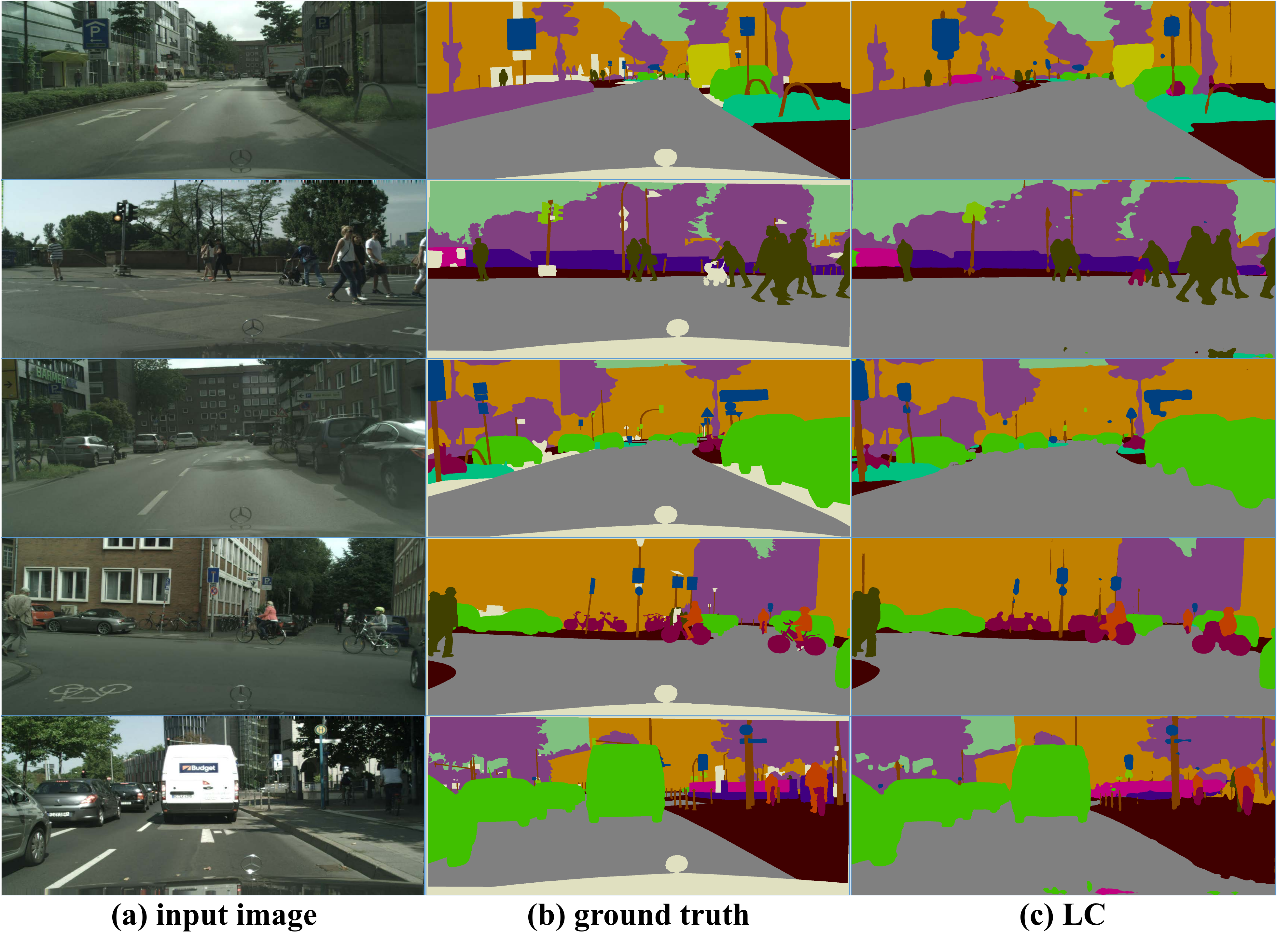}
	\caption{Visual quality of LC label maps: (a) input image (b) ground truth (white labels indicating ambiguous regions) and (c) LC.}
	\label{fig:visual_cityscape}
\end{figure*}

\subsection{Visual Quality Comparison}

We inspect visual quality of obtained label maps on VOC12 \textit{validation set}.
Fig.~\ref{fig:visual} demonstrates the comparisons of LC with DPN \cite{liu2015semantic} and DeepLab-v2 \cite{CP2016Deeplab}.
We use the publicly released model to re-generate label maps of DeepLab-v2 while the results of DPN are downloaded from their project page.
LC generally makes more accurate predictions.
We also include more examples of LC label maps on Cityscapes dataset \cite{Cordts2016Cityscapes} in Fig. \ref{fig:visual_cityscape}.

%% file: conclusion_v3_ccloy.tex
Deep layer cascade (LC) is proposed in this work to simultaneously improve the accuracy and speed of semantic image segmentation.
It has three advantages over previous approaches.
First, LC adopts a ``difficulty-aware'' learning paradigm, where earlier stages are trained to handle easy and confident regions and hard regions are progressively forwarded to later stages.
Secondly, since each stage only processes part of the input, LC can accelerate both training and testing by the usage of region convolution.
Thirdly, LC is an end-to-end trainable framework that jointly optimizes the feature learning for different regions, thus achieving state-of-the-art performance on both PASCAL VOC and Cityscapes datasets.
LC is capable of running in real-time yet still yielding competitive accuracies.